\documentclass{article}

\usepackage[]{arxiv}

\usepackage[utf8]{inputenc} 
\usepackage[T1]{fontenc}    
\usepackage{hyperref}       
\usepackage{url}            
\usepackage{booktabs}       
\usepackage{amsfonts}       
\usepackage{nicefrac}       
\usepackage{microtype}      

\usepackage{gensymb}
\usepackage{amsmath}
\usepackage{graphicx}
\usepackage{textcomp}

\title{Feature Products Yield Efficient Networks}

\author{%
  Philipp Gr\"uning\\
  Institute for Neuro- and Bioinformatics\\
  University of L\"ubeck\\
  L\"ubeck, Germany \\
  \texttt{gruening@inb.uni-luebeck.de} \\
  \And
  Thomas Martinetz\\
  Institute for Neuro- and Bioinformatics\\
  University of L\"ubeck\\
  L\"ubeck, Germany \\
  \texttt{martinet@inb.uni-luebeck.de} \\
  \And
  Erhardt Barth\\
  Institute for Neuro- and Bioinformatics\\
  University of L\"ubeck\\
  L\"ubeck, Germany \\
  \texttt{barth@inb.uni-luebeck.de} \\
}

\begin{document}

\maketitle

\begin{abstract}
  We introduce Feature-Product networks (FP-nets) as a novel deep-network architecture based on a new building block inspired by principles of biological vision. For each input feature map, a so-called FP-block learns two different filters, the outputs of which are then multiplied. Such FP-blocks are inspired by models of end-stopped neurons, which are common in cortical areas V1 and especially in V2. Convolutional neural networks can be transformed into parameter-efficient FP-nets by substituting conventional blocks of regular convolutions with FP-blocks. In this way, we create several novel FP-nets based on state-of-the-art networks and evaluate them on the Cifar-10 and ImageNet challenges. We show that the use of FP-blocks reduces the number of parameters significantly without decreasing generalization capability. Since so far heuristics and search algorithms have been used to find more efficient networks, it seems remarkable that we can obtain even more efficient networks based on a novel bio-inspired design principle.
\end{abstract}

\section{Introduction}
Deep learning has been boosted by the principle of 'deeper is better' but meanwhile computational efficiency becomes an issue and a plethora of works propose efficient model architectures. Most often, heuristics and architecture search algorithms are used to reduce the number of parameters. An alternative approach used here is to incorporate further principles from human vision. \\
Traditionally, convolutional neural networks (CNNs) had been designed to mimic orientation-selective simple- and complex cells of the primary visual cortex (V1). Accordingly, CNNs learn linear filters that reduce the entropy of natural images by encoding oriented straight patterns (1D regions) such as vertical- and horizontal edges.
In cortical area V2, however, the majority of cells are end-stopped to different degrees~\cite{hubel1965receptive}. End-stopped cells are thought to detect 2D regions such as junctions and corners. Since 2D regions are unique and sparse in natural images~\cite{MoBa00,BaWa2000,ZeBaWe93a} they have the potential of representing images efficiently.\\
A standard way of modeling end-stopped cells is to multiply outputs of orientation-selective cells~\cite{ZeBa90a}, resulting in an AND-combination of simple-cell outputs. For example, a corner can be detected by the logical combination of 'horizontal edge AND vertical edge.'
Arguably, deep CNNs have the capability of modeling such behavior with extensive successions of convolution-ReLU layers. However, explicit modeling of end-stopped cells should result in more efficient networks, since the required AND operations are introduced directly.\\
Following this idea, we present \textit{FP-nets}: CNN architectures where in strategic positions standard components are replaced with our novel \textit{FP-block}. Inspired by how end-stopped cells function, the FP-block learns two filters for an input feature map and multiplies them. This can be done efficiently by processing one input tensor with two different depth-wise separable convolutions (dws-convolutions) in parallel. Subsequently, the multiplication results are recombined by a $1 \times 1$ convolution. 

\subsection{Related Work}
With FP-nets, we aim to create more efficient architectures. An important step in that direction was the use of dws-convolutions as used with the MobileNet~\cite{mobilenet} architecture: regular convolutions are replaced with channel-wise filter operations and subsequent $1 \times 1$ convolutions to ensure information flow between different channels. Inverted bottleneck residual blocks~\cite{mobilenetv2} further improved the results: in addition to residual connections known from the ResNet
~\cite{resnet}, they first expand the number of feature maps, apply dws-convolutions, and then shrink the result to the desired output width. This structure, together with the squeeze-and-excitation method~\cite{squeeze_and_excitation}, is also mainly employed in the EfficientNet~\cite{efficientnet}, which shows remarkable results while saving parameters. Shuffled group convolutions~\cite{shufflenet} in the $1 \times 1$ recombination layers can reduce the number of parameters even further. Weight pruning~\cite{lottery_weight_pruning} is an alternative that removes unnecessary weights without altering the network structure directly.\\
In our work, we address multiplications as a key component. With CNNs, multiplications are also used in re-weighting channel distributions~\cite{squeeze_and_excitation,eca_net}. Furthermore, attention based models focus on specific locations by multiplying an attention map to feature-maps~\cite{attention_unet}.\\
With Selective Kernel Networks, Li et al.~\cite{selective_kernel_net} present a bio-inspired architecture, where neurons are able to adjust their receptive field size based on the input. We consider Zoumpourlis et al.'s~\cite{zoumpourlis2017non} work on non-linear convolution filters to be very close to our own ideas. Influenced by computational models of the visual cortex, they show that altering the first layer of a CNN to incorporate quadratic forms through the Volterra Kernel~\cite{volterra} can improve generalization. The restriction to the first layer is due to a heavy computational overhead. Our FP-block can also be interpreted as a second-order Volterra kernel (see Section~\ref{sec:methods}), but has much fewer parameters and there are no constraints as to where we can place it. Before the advent of deep learning, second-order terms have been explored in related fields, e.g.~\cite{bergstra_quad,berkes_wiscott_quad}.

\section{Methods}
\label{sec:methods}
\begin{figure}[h]
  \centering
  \centerline{\includegraphics[width=4cm]{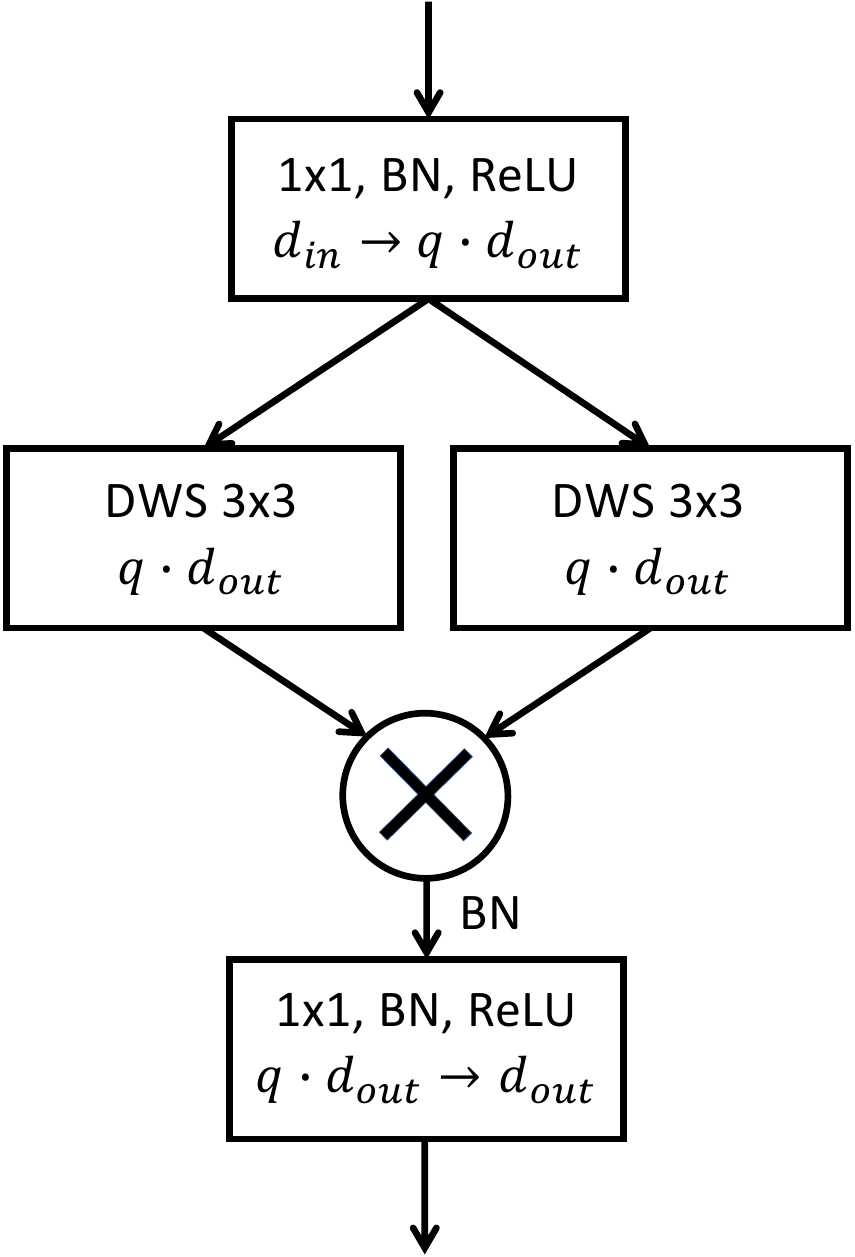}}
\caption{
FP-block: our proposed structure simulates the behaviour of end-stopped cells. For each input feature map, two filters are learned and subsequently multiplied, creating more selective outputs (e.g. corner-detection). This is efficiently achieved by using two depth-wise-separable (dws-) convolutions. This operation is enframed by two $1 \times 1$ convolutions with batch normalization (BN) and ReLU that alter the input and output dimensions $d_{in}$ and $d_{out}$.
}
\label{fig:fp_block}
\end{figure}

FP-nets are inspired by a principle of biological vision: end-stopped cells in the primary cortical areas provide an efficient visual representation by suppressing 1D features and are known to require AND-type non-linearities~\cite{ZeBa90a}. Arguably, large sequences of convolution layers with ReLUs are capable of providing such non-linearities. Nonetheless, modeling end-stopped cells more explicitly by introducing FP-blocks should yield more efficient architectures. Our computationally feasible approach is to create a structure that multiplies outputs of dws-convolutions. More specifically, for each input feature-map, two filters are learned and subsequently multiplied. Accordingly, \textit{filter-pairs} can emerge that can suppress specific (1D) signals, an encoding that a single filter could not perform. For example, a horizontal and vertical filter pair (e.g., the Sobel-x and Sobel-y filter) can suppress straight edges and focus on corners. 
Let's regard a single patch of an input feature map and a filter-pair $\vec{x},(\vec{f_{a}},\vec{f_{b}}) \in \mathbb{R}^{k^2}$. Essentially, our approach computes
\begin{align}
    g(\vec{x}) &= (\vec{f_{a}}\vec{x})(\vec{f_{b}}\vec{x})\\
    &= (\sum_{ i=0 }^{ n }{ f_{a}^i x^i })(\sum_{ j=0 }^{ n }{ f_{b}^j x^j }) \\
    &= \sum_{ i=0 }^{ n }{\sum_{ j=0 }^{ n }{ w^{ij} x^j x^i}}, 
\end{align}
with $n = k^2 -1$. This shows that signals that are orthogonal to any of the two filters are entirely suppressed. Accordingly, a more selective output can be produced compared to a single filter operation. Furthermore, we see that the output is a weighted sum of all possible $k^4$ pixel pairs. Hence, it can be regarded as a second-order Volterra kernel. However, the above approach is more parameter-efficient since the $k^4$ weights are derived from all pairs of the $2k^2$ filter weights.\\
After multiplication, batch normalization (without affine rescaling) is used and it reduces the risk of exploding values for $g(\vec{x})$. The above-described operations are, of course, done in parallel for several input feature maps. Accordingly, a subsequent $1 \times 1$ convolution layer with batch normalization and ReLU recombines the different outputs and enables cross-feature-map information flow. A design choice that is best known from the MobileNet~\cite{mobilenet} architecture.\\
Finally, we designed the main building block of FP-nets, the \textit{FP-block}, as follows: an input tensor is first processed by a $1 \times 1$ convolution with batch normalization and ReLU that changes the width of the input tensor $d_{in}$ to $q \cdot d_{out}$; where $d_{out}$ is the desired output dimension of the whole block and $q$ is an expansion factor. Subsequently, we compute the outputs of two parallel dws-convolutions, multiply and normalize them. Afterwards, the result is recombined to a tensor with dimension $d_{out}$ by a $1 \times 1$ convolution with batch normalization and ReLU. Figure \ref{fig:fp_block} shows a sketch of the FP-block's structure.\\
Considering an input tensor with $d_{in}$ feature maps, the number of parameters for an FP-block is:
\begin{equation}
    N = q d_{in} d_{out} + 2 q k^2  d_{out} + q d_{out}^2.
\end{equation}
As known for the inverted bottleneck residual blocks~\cite{mobilenetv2}, we found that a higher number of feature maps (a higher $q$) was beneficial when using dws-convolutions.

\section{Experiments}
We assume that the human-vision-inspired FP-block is beneficial for neural network architectures in computer vision. An FP-net's efficiency may be higher because it allows for a more explicit modeling of AND-terms that, for example, can detect pairs of specific orientations. Accordingly, a less deep FP-net may perform just as well as a deeper standard architecture. To test this hypothesis, we created several experiments based on Cifar-10~\cite{cifar10} and ImageNet~\cite{imagenet} with FP-nets that were adapted versions of state-of-the-art networks. To describe the adaptations we performed, we adopt the nomenclature of the ResNet paper~\cite{resnet}: \textit{layers} are higher-order structures in an architecture that contains a number of \textit{blocks}. In the following experiments we often replaced the original layers with \textit{FP-layers} that contained one or several \textit{FP-blocks} (see Section \ref{sec:methods}). We call a CNN architecture that incorporates FP-blocks an \textit{FP}-network. If a specific architecture is used as the basis for the FP-net, e.g. the ResNet, we refer to it as \textit{FP-ResNet}.\\
For several ResNet architectures, we segmented layers containing several sequences of convolution and non-linearities (blocks). We replaced those layers with FP-layers consisting of fewer FP-blocks that effectively reduced the resulting depth and reduced the number of parameters. By comparing several of those configurations with the original architectures, we could determine where to place the FP-layers in order to decrease the number of parameters without reducing the generalization capability. In the remainder of the paper, we will refer to specific configurations with binary strings. For example, configuration '0101' will mean that we replaced the second and fourth original layer with an FP-layer and we left the first and third layer as is.

\subsection{FP-ResNets on Cifar-10}
\label{sec:exp_resnet_cifar}
The original ResNet paper presents results of differently-sized ResNet architectures that were specifically designed for Cifar-10.
These models, that vary in depth, are already parameter-efficient with a viable test error ranging from $9\%$ to $6\%$ and parameters ranging from 270K to 660K. We aimed to save even more parameters. Each original model starts with a $3 \times 3$ convolution with batch normalization and ReLU. Subsequently, three layers containing $n$ \textit{basic blocks} are stacked onto each other. The second and the third layer's initial basic block starts with a $3 \times 3$ convolution with stride 2. A basic block contains two sequences of $3 \times 3$ convolution, batch normalization, and ReLU. Additionally, each basic block employs the residual shortcut connection giving the ResNet its name. We closely examined the following models: ResNet-20, ResNet-32, ResNet-44 with $n$ being 3, 5, and 7 respectively. Note that in the ResNet paper, the term 'first layer' denotes the first convolution \textit{with} the subsequent $n$ basic blocks. In this work, the term first layer is defined as \textit{only} the $n$ basic blocks; thus, our configurations do not alter the first convolution.\\
Our modified architectures replaced one or two specific layers of each ResNet architecture with $m$ FP-blocks. For ResNet-20, ResNet-32, ResNet-44, $m$ was 1, 3, and 5 respectively: the number of blocks taken from the next-smaller model.
As described in Section \ref{sec:methods}, the FP-block was a sequence, starting with a $1 \times 1$ convolution with batch normalization and ReLU, that mapped the input dimension $d_{in}$ to $2d_{out}$ (expansion factor $q=2$). Subsequently, we computed the depth-wise filter-pairs with two parallel dws-convolutions. By using another $1 \times 1$ convolution with batch normalization and ReLU, the $2d_{out}$ feature maps were mapped to $d_{out}$ feature maps. Instead of stride 2 convolutions, we appended a max-pooling operation to the first block. Note that, we did not use shortcut connections.\\
Replacing one or more of the original layers, yielded $2^3 - 1 = 7$ possible configurations. We disregarded the configuration '111' because it did not produce feasible results when using FP-blocks. Hence, we trained six configurations. We report the minimal test error on Cifar-10 averaged over four or eight runs with different seeds (different weight initialization and batch-order). We trained each model with the same hyperparameters: 200 epochs, with stochastic gradient descent, an initial learning rate of 0.1 that we reduced after 100 epochs to 0.01 and after 150 epochs to 0.001. We used a momentum of 0.9 and a batch size of 128. These parameters and all data augmentation operations were similar to those used in the Github repository of Idelbayev~\cite{c10_resnet_github}. On average, training a model took 50 minutes on a Nvidia RTX2080.\\

\subsection{FP-ResNet on ImageNet}
\label{sec:exp_resnet_imagenet}
To test our design choices on a more challenging dataset, we trained an FP-ResNet on the ImageNet benchmark. Our network was based on the ResNet-50 architecture containing four layers made of \textit{bottleneck blocks}. Initial tests showed that replacing the second and fourth layer was a viable choice. For each, we used one FP-block with an expansion factor $q=1$. Note that, we again used no shortcut connections. We employed the same data augmentation and hyperparameters as described in the ImageNet training example of the Pytorch repository~\cite{pytorch_imagenet_example}. We report the validation error of the last epoch and compare our result to the validation errors of Pytorch's~\cite{pytorch} pre-trained ResNets. We trained the model on four Nvidia RTX 2080 for 64 hours.


\section{Results}

\begin{figure}[h]
  \centering
  \centerline{\includegraphics[width=12cm]{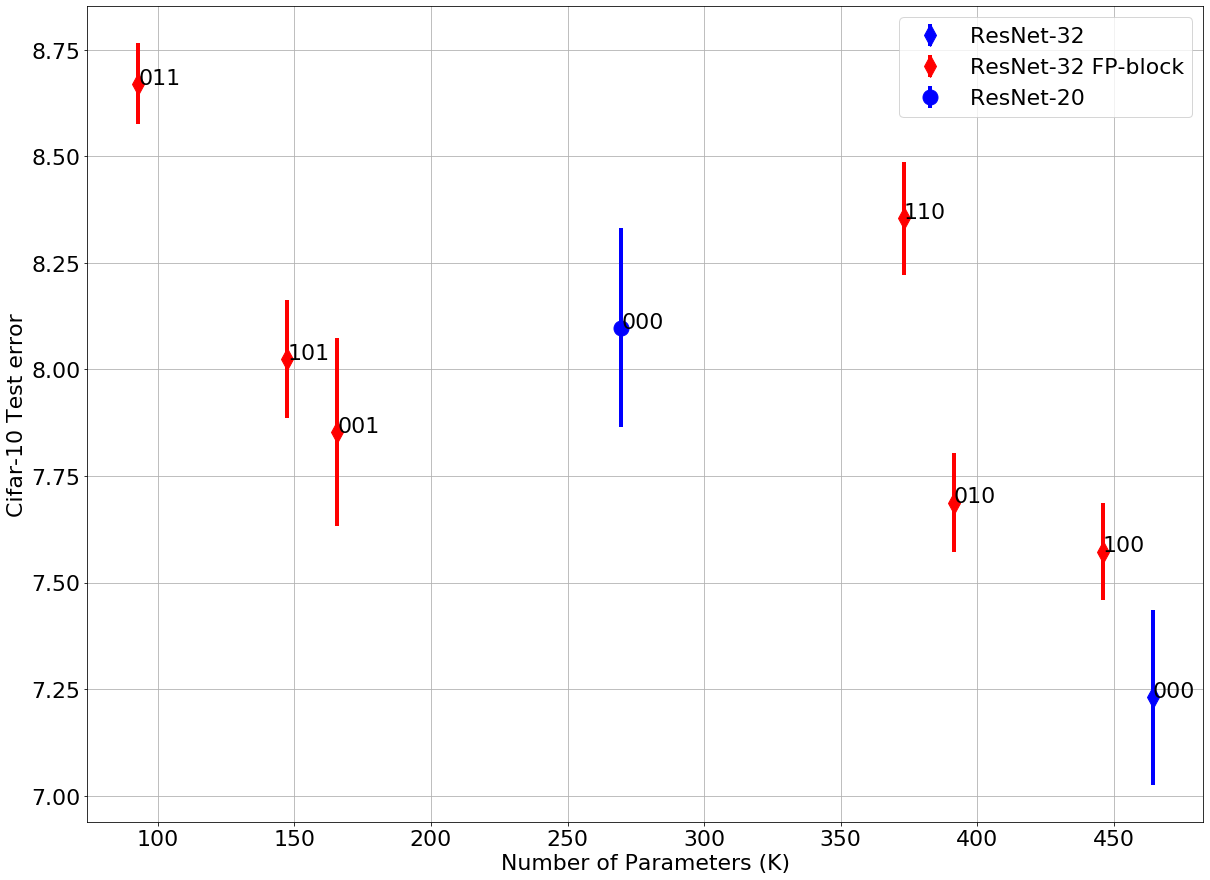}}
\caption{
Results averaged over at least 4 runs with standard deviation for layer substitutions of the ResNet-32 on Cifar-10:
the original ResNet-32 is made of three layers that each contain five basic blocks. The binary string next to the point indicates which layer was substituted. E.g., for '101' we substituted the first and third original layer with FP-layers, each consisting of three FP-blocks. Changing the third layer reduced the number of parameters considerably, while the error increased only slightly.
}
\label{fig:resnet_32}
\end{figure}

\begin{figure}[h]
  \centering
  \centerline{\includegraphics[width=12cm]{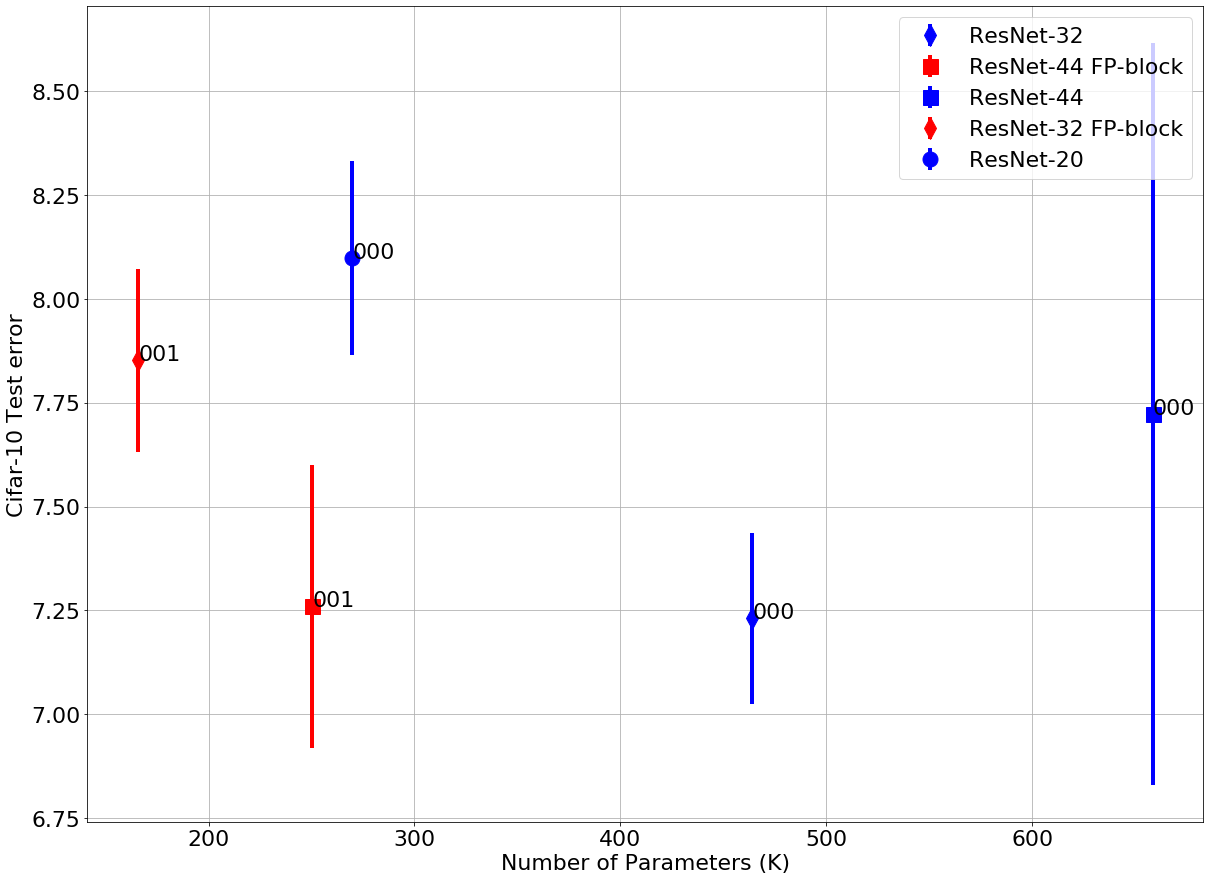}}
\caption{
Results averaged over at least 4 runs with standard deviation for substituting the last layer on different ResNet architectures: the blue color shows the ResNet models (point: ResNet-20, diamond: ResNet-32, square: ResNet-44). The red color shows the FP-nets; where we substituted the original last layer that contains $n$ basic blocks (7 for the ResNet-44, 5 for the ResNet-32, 3 for the ResNet-20) with fewer FP-blocks (5 for the FP-ResNet-44, 3 for the FP-ResNet-32, 1 for the FP-ResNet-20). Downsizing layers at specific locations could significantly reduce the number of parameters with only a small decrease in generalization capability.
}
\label{fig:resnet_00x}
\end{figure}

Our main result is that by using FP-blocks in state-of-the-art networks we obtained a number of parameter-efficient FP-nets.\\
Figure \ref{fig:resnet_32} shows the results for six different FP-ResNets-32, and the baselines ResNet-32 and ResNet-20. The choice of which layer to replace had a high impact on the number of parameters because the feature-map dimensions increased with growing depth. Accordingly, substituting the middle and last layer decreased the number of parameters more than substituting the first layer. FP-nets of type '001' that had an FP-layer on the last position showed a good trade-off between parameter-efficiency and generalization. Thus, we could create FP-nets, with the ResNet-32 as base network, that had much fewer parameters than the ResNet-20 and yet performed better on the test set. Overall, the configuration '001' was most efficient.
Therefore, we present further results for configuration '001' with different base-networks - see Figure \ref{fig:resnet_00x}. Here, an even better network emerged: using the ResNet-44 as base network, we created an FP-net with performance comparable to the ResNet-32. However, we still managed to have fewer parameters than the ResNet-20.
Our data show a pattern: deeper networks with FP-layers (FP-nets) outperform shallower baseline networks although the FP-nets have less parameters.\\
Remarkably, we could replicate this pattern on ImageNet. We present the results in Table \ref{tab:imagenet}: our FP-ResNet-50 achieves a validation error comparable to the original ResNet-50. Moreover, it outperforms the ResNet-34 and ResNet-18, and has 5M parameters less than the ResNet-34. Furthermore, our hypothesis was confirmed that a shallower FP-net (the FP-ResNet has fewer layers than the ResNet-50) can compete with a deeper standard architecture.\\
At last, for the ResNet-32, we conducted an ablation study to quantify the effect of the filter-pairs apart from other design decisions: namely, using dws-convolutions that were enframed with $1 \times 1$ convolutions. Instead of computing two dws-convolutions, we only computed one and used a ReLU as non-linearity. The results are shown in Table \ref{tab:ablation}. For the ResNet-32, using an FP-layer decreased the average test error by $0.4\%$. For configuration '001' on the FP-ResNet-32, 3456 parameters were added. Accordingly, when using FP-blocks, we were able to improve generalization over simple dws-convolutions with only a few more parameters.


\begin{table}[]
\caption{
Ablation study: to show that multiplying depth-wise separable filter pairs has an actual beneficial effect, we compared our results to a single depth-wise separable filter output with an additional ReLU. 
}
\label{tab:ablation}
\centering
\begin{tabular}{lll}
\hline
Type                               &N. Param. (K) & Val. Error                      \\ \hline
ResNet-32 001 Single Filter ReLU   &162           &8.25 $\pm$ 0.17\\
ResNet-32 001 Filter Pairs         &166           &7.85 $\pm$ 0.22\\  \hline
\end{tabular}
\end{table}

\begin{table}[]
\caption{
ImageNet validation results: we compared our architecture to the Pytorch~\cite{pytorch} pre-trained ResNets.
}
\label{tab:imagenet}
\centering
\begin{tabular}{lll}
\hline
Type        &N. Param. (M) & Val. Error \\ \hline
ResNet-18   &11            &30.34 \\
ResNet-34   &21            &26.78 \\
ResNet-50   &23            &23.88 \\
ours        &16            &23.87 \\ \hline
\end{tabular}
\end{table}

\section{Discussion}
Convolutional neural networks are the state-of-the-art solution for most computer vision tasks. One reason these networks perform so well may be that they incorporate properties inspired by biological vision systems, which through millions of years of evolution have adapted to the given problem structure. Simple cortical cells can be modeled with convolutions, i.e. a sparse matrix multiplication with many similar coefficients. This weight sharing decreases the search space for network parameters significantly over standard multi-layer perceptrons. The design paradigm is clear: find and model behaviour from biological systems to create neural networks that are more suitable for a specific task. In this work, we argued that principles from biological vision can be further exploited. To this end,  we introduced the FP-block as a structure that simulates the behavior of end-stopped cells. Our results show that by adding this structure to established architectures one can indeed create FP-nets, as hybrid CNNs, that are more efficient, i.e., they generalize well with less parameters. 
Moreover, transforming any standard CNN to an FP-net is straightforward: we did not change the structures of the base networks and did not perform any hyperparameter tuning. In conclusion, we can recommend using FP-blocks as building blocks for more efficient deep networks, the FP-nets.

\section*{Broader Impact}
Who may benefit from this research?\\
The short answer: (i) all applications that require efficient networks, and (ii) research on deep learning inspired by biological vision.\\
Our first main point is that drawing inspiration from biological vision systems is a viable approach. Hopefully, this synergy of neuroscience and engineering will lead to further insights in both fields. Moreover, we presented a new practical building block for parameter-efficient networks.  FP-blocks may become an essential part of a new 'MobileNet' or 'EfficientNet.' 
Apart from the researcher's perspective, efficient machine learning has an even broader impact. First of all, it can help with smart devices aimed to make everyday life easier. Furthermore, it democratizes AI-research: if the newest and best neural network can only be trained on a large GPU cluster, only a few people will adopt machine learning in new areas where it might be beneficial. Finally, smaller models consume less energy.
b) - c): Not applicable.
\bibliographystyle{ieeetr}
\bibliography{literature}
\end{document}